\documentclass{article}

\usepackage[preprint]{neurips_2022}
\usepackage{multirow}
\usepackage{multicol}
\usepackage{makecell}
\usepackage{booktabs}
\usepackage[utf8]{inputenc} % allow utf-8 input
\usepackage[T1]{fontenc}    % use 8-bit T1 fonts
\usepackage{hyperref}       % hyperlinks
\usepackage{url}            % simple URL typesetting
\usepackage{booktabs}       % professional-quality tables
\usepackage{amsfonts}       % blackboard math symbols
\usepackage{nicefrac}       % compact symbols for 1/2, etc.
\usepackage{microtype}      % microtypography
\usepackage{xcolor}         % colors
\newcommand{\modelname}{{\usefont{T1}{ppl}{m}{n}SchemaPro}}

\usepackage{multicol}
\usepackage{bigstrut}
\usepackage{graphicx}

\usepackage{soul}
\usepackage{tablefootnote}
\usepackage[normalem]{ulem}
\usepackage{multirow}
\usepackage{makecell}
\usepackage{enumitem}
\usepackage{adjustbox}
\usepackage{amsmath}
\usepackage{array}
\usepackage{graphicx}
\usepackage{comment}
\usepackage{booktabs}
\usepackage{xcolor,colortbl} 
\usepackage{algorithm}

\usepackage{bbm}
\usepackage{tabularx}
\usepackage{hyperref}

\usepackage{graphicx}
\usepackage{subcaption}
\usepackage{arydshln}
\usepackage{array}
\usepackage{booktabs}
\usepackage{pifont}
\usepackage{cleveref}
\crefname{section}{§}{§§}
\usepackage{amsthm}
\usepackage{todonotes}
\usepackage{color}
\usepackage{amsmath, bm, amsfonts}

\title{{Improving Task Generalization via \\ Unified Schema Prompt}}

\author{Wanjun Zhong$^{1}$, Yifan Gao$^{4}$, Ning Ding$^3$, Zhiyuan Liu$^3$, \\\textbf{Ming Zhou}$^5$, \textbf{Jiahai Wang$^1$, Jian Yin$^1$ and Nan Duan$^2$} \\
	$^1$ Sun Yat-sen University \quad $^2$ Microsoft Research Asia \\
    $^3$ Tsinghua University \quad $^4$ Chinese University of Hong Kong \quad $^5$ Langboat Technology  \\
	{\tt \{zhongwj25@mail2, wangjiah@mail,issjyin@mail\}.sysu.edu.cn}\\
	{\tt yfgao@cse.cuhk.edu.hk; \tt liuzy@tsinghua.edu.cn} \\
	{\tt \{dingn18\}@mails.tsinghua.edu.cn} \\
	{\tt nanduan@microsoft.com}; \tt zhouming@chuangxin.com\\ 
}

\begin{document}

\maketitle

\begin{abstract}
  Task generalization has been a long-standing challenge in Natural Language Processing (NLP). 
  Recent research attempts to improve the task generalization ability of pre-trained language models by mapping NLP tasks into human-readable prompted forms. 
  However, these approaches require laborious and inflexible manual collection of prompts, and different prompts on the same downstream task may receive unstable performance. We propose Unified Schema Prompt, a flexible and extensible prompting method, which automatically customizes the learnable prompts for each task according to the task input schema. It models the shared knowledge between tasks, while keeping the characteristics of different task schema, and thus enhances task generalization ability. 
  The schema prompt takes the explicit data structure of each task to formulate prompts so that little human effort is involved.
  To test the task generalization ability of schema prompt at scale, we conduct schema prompt-based multitask pre-training on a wide variety of general NLP tasks. The framework achieves strong zero-shot and few-shot generalization performance on 16 unseen downstream tasks from 8 task types (e.g., QA, NLI, etc). Furthermore, comprehensive analyses demonstrate the effectiveness of each component in the schema prompt, its flexibility in task compositionality, and its ability to improve performance under a full-data fine-tuning setting.
\end{abstract}

\section{Introduction}

Task generalization can be viewed as out-of-domain adaptation to unseen tasks with diverse task-specific knowledge.
% and input-output formats. 
% It is extremely challenging especially before the large language model was invented, as it always requires sufficient supervised data to achieve better performance. 
% Pre-trained language models \citep{devlin2018bert} unify a wide range of tasks in the same architecture to share the cross-task knowledge as much as possible, but still require task-specific dataset for fine-tuning.
Pre-trained language models \citep{devlin2018bert} achieve state-of-the-art performance on a wide range of tasks without substantial task-specific architecture modifications, but they still require fine-tuning with one additional layer on task-specific datasets \citep{li2020unsupervised, gao2020discern, sun2022unified}.
% Recent research has shown that prompting in large language models exhibit the ability of task generalization \citep{brown2020language} with less supervisions, which indicates that there may be tight connections and shared common knowledge between various tasks. 
Moreover, fine-tuned pre-trained model on a specific dataset cannot be generalized to other unseen tasks, especially when different tasks have diverse formats of inputs and outputs.
% Moreover, the traditional way of fine-tuning with task-specific objectives is impractical for further task generalization because different tasks have diverse formats of inputs and outputs. 
This observation promotes recent works \citep{sanh2021multitask,raffel2020exploring,xu2022zeroprompt} on using prompting methods to improve task generalization through explicit multi-task learning.
% , in which the inputs and outputs of different tasks are reformulated into a unified format.
% and solving various tasks with a unified encoder-decoder framework, to model their commonality and improve task generalization ability.
% It has been proved that the prompting-based unified paradigm significantly improves zero-shot task generalization ability \citep{sanh2021multitask}.
% a unified encoder-decoder paradigm to solve various tasks within a same framework to model their commonality and improve task generalization ability. 
% These works typically formulate the input using human-written natural language (NL) prompts as a prior.
Formulating the input of different tasks using human-written natural language (NL) prompts into a unified format, the prompting-based unified paradigm significantly improves zero-shot task generalization ability \citep{sanh2021multitask}.

Despite its great success, the paradigm of writing NL prompts involves huge human efforts. For example, T0 \citep{sanh2021multitask} relies on a crowd-sourcing platform to collect 1939 human-written NL prompts from 36 contributors.
These human-written prompts are tied with their tasks so that they are infeasible to be generalized to new tasks.
Moreover, human-written prompts on the same task usually receive performance variance because they are not equivalently accurate in task description.
% The research question behind this paper is -- \textit{''Is there any prompting approach exempting from manually writing prompts while keeping the generalization ability?''}.
Under the curiosity of exempting from manually writing prompts while keeping the generalization ability, we find the explicit data schema in the datasets of many NLP tasks can be used as-is for automatic prompting.
For example, a QA task has the input schema ``{\textit{question}: ...; \textit{answer}: ...; \textit{passage}: ...}''. 
Instead of writing a prompt like ``What is the best answer to the question ... given the passage ...?'', the compositional input components in the schema have already provided a super informative way for prompting: ``question: xxx, passage: xxx, answer:?''.
In addition to alleviating the manual involvement for prompt writing, keeping the original data schema for prompting brings two benefits:
(1) Treating task schema as keys in prompting can model different combinations of input components to discriminate different tasks. 
For example, \textit{natural language inference (NLI)} task has input components ``(\textit{premise, hypothesis})'', and \textit{question answering (QA)} task has ``\textit{(passage,question)}''.
(2) Different tasks may have shared input schema so that schema-specific knowledge could be shared across tasks.
For example, schemas of QA tasks share common component like ``\textit{question}'', and the tasks of \textit{summarization} and \textit{topic classification} have common input ``\textit{passage}''. 

\begin{figure*}[t]
	\centering
	\includegraphics[width=\textwidth]{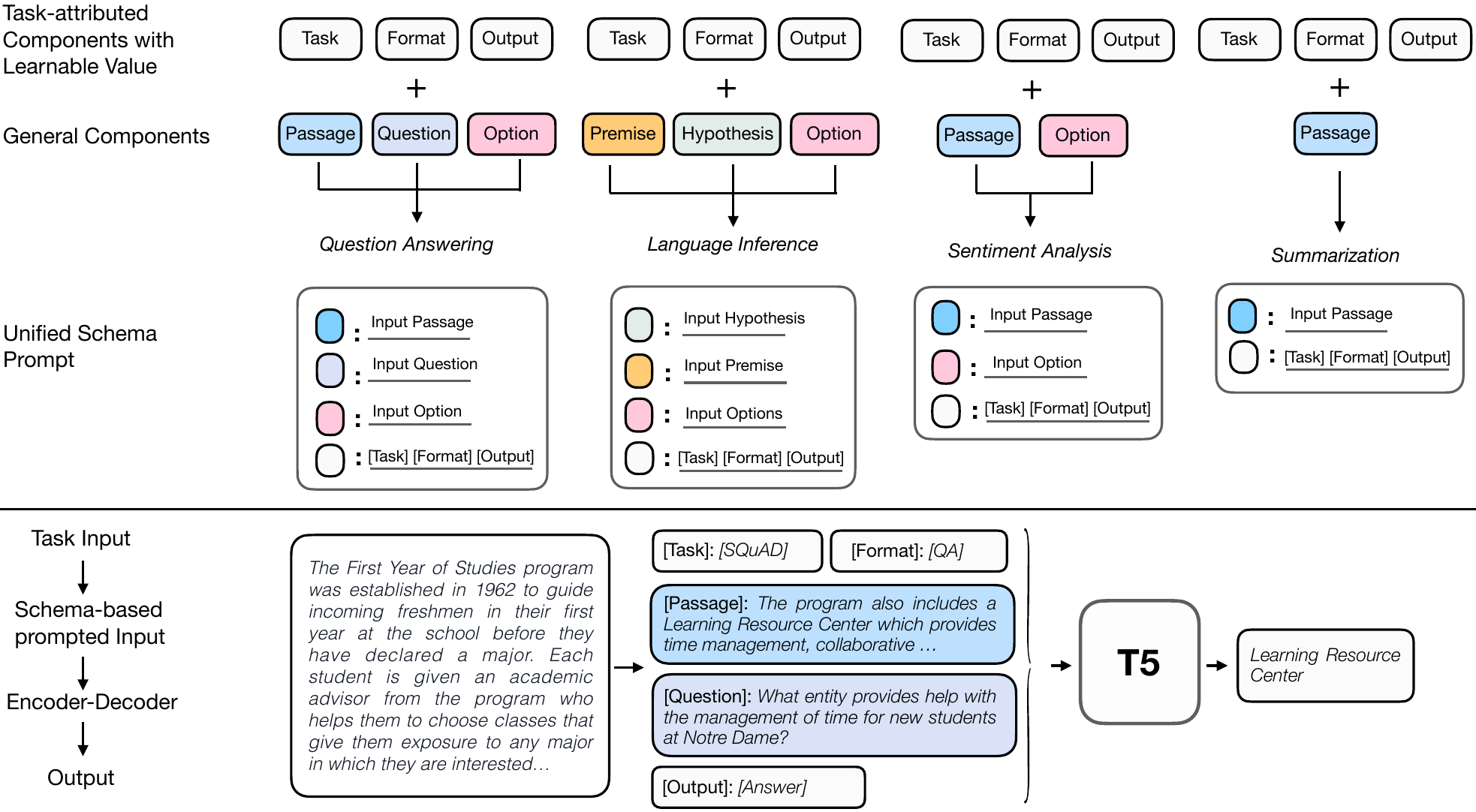}
	\caption{The overall framework of \modelname, which automatically composes learnable and shareable prompts for tasks according to their task schemas, and organizes the inputs of every NLP tasks into a unified format.
% 	organizes the inputs of every NLP tasks into a unified format with learnable and shareable prompts.
	Each colored box indicates a specific component type like ``\textit{Question}'', and is represented with key prompts. 
	The white box indicates special components (i.e., \textit{Format}, \textit{Task}, \textit{Output}) representing tasks attributes.
	Importantly, the representations of every component types and task-attributed values residing in the special components are all learnable and storable.
	Each element in square brackets or colored boxes is a specific group of learnable soft prompts.}
% 	For example, a QA task input can be formatted as: ``\{\texttt{[Passage]:} Super Bowl 50 ...  ;\texttt{[Question]:} Which NFL team ... ? ; \texttt{[Format]}:\texttt{[QA]}; \texttt{[Task]}:\texttt{[SQuAD]}; \texttt{[Output]}:\texttt{[Answer]}\}''. }
% 			We omit the fused evidence block from the figure for simplification.}
	\label{fig:framework}
\end{figure*}

With the aforementioned motivations, we propose a task schema-based prompting method, \textbf{\modelname}, to simultaneously model the shared knowledge and variances between wide variety of tasks, and to alleviate human efforts in prompt writing.
% , and more importantly
% , to simultaneously model the shared knowledge and variances between wide variety of tasks. 
As shown in Fig.~\ref{fig:framework}, the schema prompt is composed of multiple components (represented as \textit{key-value} pairs), whose composition is defined by the task input schema itself.
The components have two types: (1) general components defined by task input schema (e.g., \textit{\{passage\}} in \textit{Summarization} task and \textit{\{premise, hypothesis\}} in \textit{NLI}), 
and (2) learnable task-specific attributes (i.e., \textit{\{task: a specific dataset\}}, \textit{\{format: a general class of a NLP task, like NLI\}}, \textit{\{output: expected output type, like answer\}}).)
In each component, the component type is a general key (e.g., \textit{passage, format, task}), and the specific instance belonging to this type is taken as a value. 
% Importantly, the semantic representations of all the keys and task-specific values are learnable and requires no human efforts in task description.

More specifically, \modelname\ has several important features. 
Firstly, each component key that helps the model in identifying the task schema in an explicit way, is represented by a group of learnable soft key prompts.
Secondly, to automatically learn the description of tasks, the values that belong to task-attributed components (i.e., \textit{task, format, output}) are also learnable continuous soft prompts. 
This design provides a discriminating capability to different task schema, plug-in flexibility of task-attributed prompts, which leads to better generalization ability and minimal manual efforts for task description.
% in which each common input component (e.g., ``\textit{passage}'',``\textit{options}'') or special learnable task attribute (e.g., ``\textit{format}'', ``\textit{output type}'') is formulated as a key, and value is the specific instance of this component. 
% More specifically, \modelname\ has several important characteristics: 
% (1) Each key that helps the model in identifying the components of various tasks in an explicit way, is represented by a learnable soft key indicator.
% (2) To specialize characteristics of tasks, \modelname\ appends several learnable task-specific {key-value} pairs (i.e., ``\textit{\{task: a specific dataset\}}'',``\textit{\{format: a general class of a NLP task, like QA\}''}).
% to specialize characteristics of tasks. 
% For the special task-specific key-value pairs, both the key indicator and the value are all learnable soft prompts, 
Moreover, the framework is extensible and flexible when a new task schema is involved -- \modelname\ only requires adding and learning a new component, or adding a new task-attributed value. 
The extensibility and flexibility bring faster model adaptation. 

We explore the effectiveness of the schema prompt in the scenario of multi-task learning for general NLP tasks. 
We first formulate the inputs of each task with the schema prompt, where the key prompts and task-attributed values are learnable during training. 
Then, we train the mixture of NLP tasks under an encoder-decoder model using T5 \citep{raffel2020exploring}.
Mixing a wide variety of NLP tasks helps the model in learning both the semantic meaning of the schema prompt and the common knowledge shared across tasks.

We evaluate the task generalization ability of schema prompt-enhanced model by zero-shot testing and few-shot adaptation on tasks that are unseen during training. 
From experiments on 16 unseen tasks belonging to 8 task types, we highlight the following findings: 
(1) \modelname\ outperforms NL prompt-based method 
% T0 \citep{sanh2021multitask} 
under both zero-shot testing and few-shot learning settings, indicating that the schema prompt enhances the task generalization ability to unseen tasks; 
(2) The ability of identifying different schema can benefit model adaptation to new tasks, as our method improves few-shot performance when adapting to unseen tasks with compositional schemas of other learned tasks;
(3) Eliminating task-attributed components residing in the schema prompt results in large performance drop, suggesting they model the task characteristics.
(4) \modelname\ can also benefit model learning even when there is enough supervised data for downstream tasks.

\section{Unified Schema Prompt for General NLP Tasks}

\subsection{Preliminaries: Unified Multi-task Learning Framework}
We introduce some definitions in NLP task generalization. 
Throughout this paper, we denote ``task'' as ``a dataset with specific data distribution or domain knowledge", and ``format'' as a common task type, like QA or NLI. 
For example, ``\textit{DREAM}''\citep{sun-etal-2019-dream} is a task and its corresponding format is ``\textit{Multiple Choice QA}''. 
Although the definition of ``task" is vague and has no a standard definition, there are still fundamental difference between different datasets with the same format that they emphasize different kinds of reasoning skills, domain knowledge, and data distribution. For example, \textit{commonsense QA} emphasizes reasoning over commonsense knowledge while \textit{Hotpot QA} emphasize multi-hop reasoning.  
Therefore, we largely follows popular works like GPT-3 \citep{brown2020language}, MAML \citep{finn2017model} and \citet{xu2022zeroprompt} and define ``task" as ``dataset with specific data distribution and domain knowledge".
% there are still fundamental differences between different 
% is that different datasets always emphasize different kinds of required skills, and the data instances in different datasets have very diverse distribution. 
% Thus, we largely follow \citet{brown2020language} and adopt the similar definition. 

Since different NLP tasks have diverse formats of inputs and outputs, modeling several NLP tasks within a unified model requires different tasks sharing a unified form. 
Prompting is a feasible way to reformat different NLP tasks into the same input-output format,  enabling the construction of a unified framework to solve various NLP tasks. 
For example, T0 \citep{sanh2021multitask} uses natural language prompting to reformat the input of \textit{natural language inference} task using the template ``\texttt{If \{Premise\} is true, is it also true that \{Hypothesis\}?}'' and formulate the output as an choice from options ``\texttt{\{yes, maybe, no\}}''. 
%requires paraphrase
With reformulated input-output pairs of NLP tasks via prompting, one can adopt encoder-decoder architecture with input fed to the encoder and target output produced by the decoder. 
% The model is trained to generate the target through standard maximum likelihood training. 

\subsection{Formulation of Schema Prompts}
\begin{figure*}[t!]
		\centering
		\includegraphics[width=0.9\textwidth]{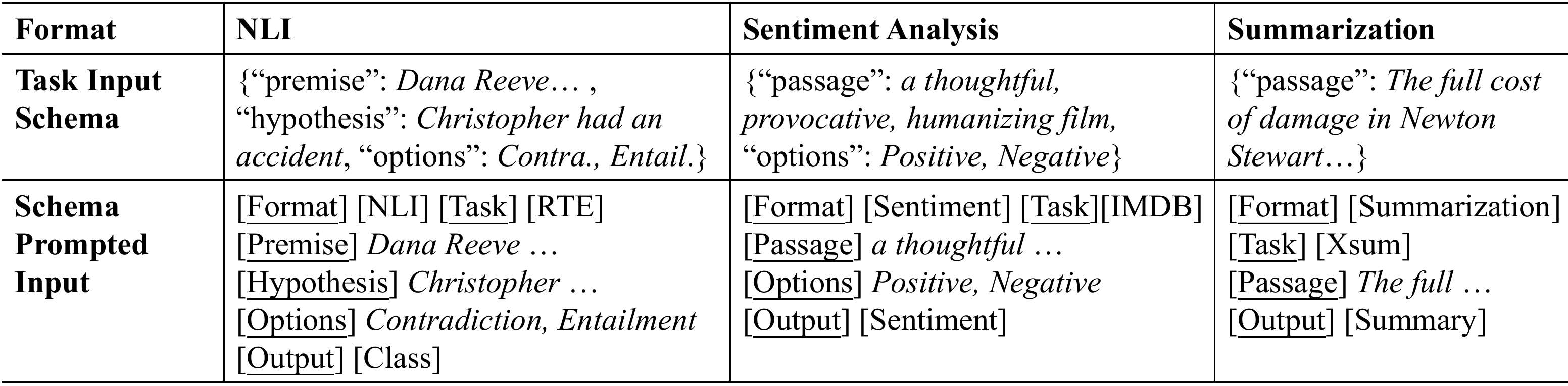}
		\caption{Examples of the task schema (json line format) and schema prompt formulated input. Items within square brackets indicates that it is a specific group of learnable soft prompts. The component keys are underlined.}
% 			We omit the fused evidence block from the figure for simplification.}
		\label{fig:example}
	\end{figure*}
% Here we detailedly introduce the definition of the schema prompt.
We design a unified task schema-based prompting method, namely \modelname, to automatically customize the prompts for each task and reformat the task inputs, involving minimal human efforts.
% We design a unified structural input schema to reformat the inputs of general NLP tasks. 
The design of \modelname\ consists of multiple components, where each component type (e.g., \textit{passage, task}) is represented as a key, and its corresponding content is taken as a value. 
For each specific task, the compositions of components are defined by the task input schema itself, and task-attributed components required for task description.
More specifically, there are two possible classes of components: (1) general components given in the task schema (e.g., \textit{passage, question, options}), where the value is a text; (2) task-attributed components used for task description (i.e., \textit{format, task, output}), where the value is a group of learnable soft prompts.
Soft prompts \citep{li-liang-2021-prefix,hambardzumyan2021warp} are learnable soft continuous embeddings, which are flexible and pluggable, and are mainly adopted for parameter-efficient model adaption for pre-trained models.

Essentially, the design of schema prompt has following specialties:
(1) To learn and customize the functionality for each component, we represent each component type with a group of learnable soft key prompts;
(2) To learn the soft task description, we also adopt learnable soft prompts as the values to represent the attributes of tasks. We define three kinds of task attributes, i.e.,  \textit{Format, Task, and Expected Output}, where \textit{Format} and \textit{Output} prompts are shareable across tasks and \textit{Task} prompts are task-specific. Fig.~\ref{fig:example} gives examples of the task schema and corresponding schema prompted input. 
Under this design, both the functionality of components and task-attributed description are learnable and storable, and the components can be dynamically composed to form the task-specific \modelname, which brings \textbf{advantages}. 
Firstly, the co-occurrence of components types, format type and output type shared across different task schemas will provide shared prior knowledge for faster and better task generalization. 
Secondly, task-specific values are pluggable and flexible to specialize the characteristics of different tasks.
Moreover, the composition of \modelname\ is automatic and dynamic, which indicates minimal required manual efforts and higher extensibility when a new task is involved as we only need to add more components or task attributed values. 

We formalize the model input. Suppose we have task A with components $C_A=\{c_1,c_2,\cdots,c_n\}$, each $c_i$ represents the $i^{th}$ component (\textit{key-value}) pair. 
Then we represent the indicator of each key as $\mathbf{k}_i$ with a group of soft key prompts. 
We represent the value $\mathbf{v}_i$ as either (1) token embeddings for value in the form of the textual content (e.g., \textit{passage: ``a thoughtful film ...''}) or (2) a group of soft value prompts for learnable task attributes (``\textit{Format}'', ``\textit{Task}'', ``\textit{Output Type}'').
 Afterwards, we represent each component $c_i$ as $\mathbf{c}_i=[\mathbf{k}_i;\mathbf{v}_i]$, which is the concatenation of $\mathbf{k}_i$ and $\mathbf{v}_i$. 
 Finally, we concatenate all the $\mathbf{c}_i$ as the reformatted model input $\mathbf{X} = [\mathbf{c}_0;\mathbf{c}_1;...;\mathbf{c}_n]$. Noting that both the key indicators and special task-attributed values are learnable, pluggable, and storable soft prompts. 
\subsection{Task Generalization}\label{sec:task_generalization}
\begin{figure*}[t]
		\centering
		\includegraphics[width=0.8\textwidth]{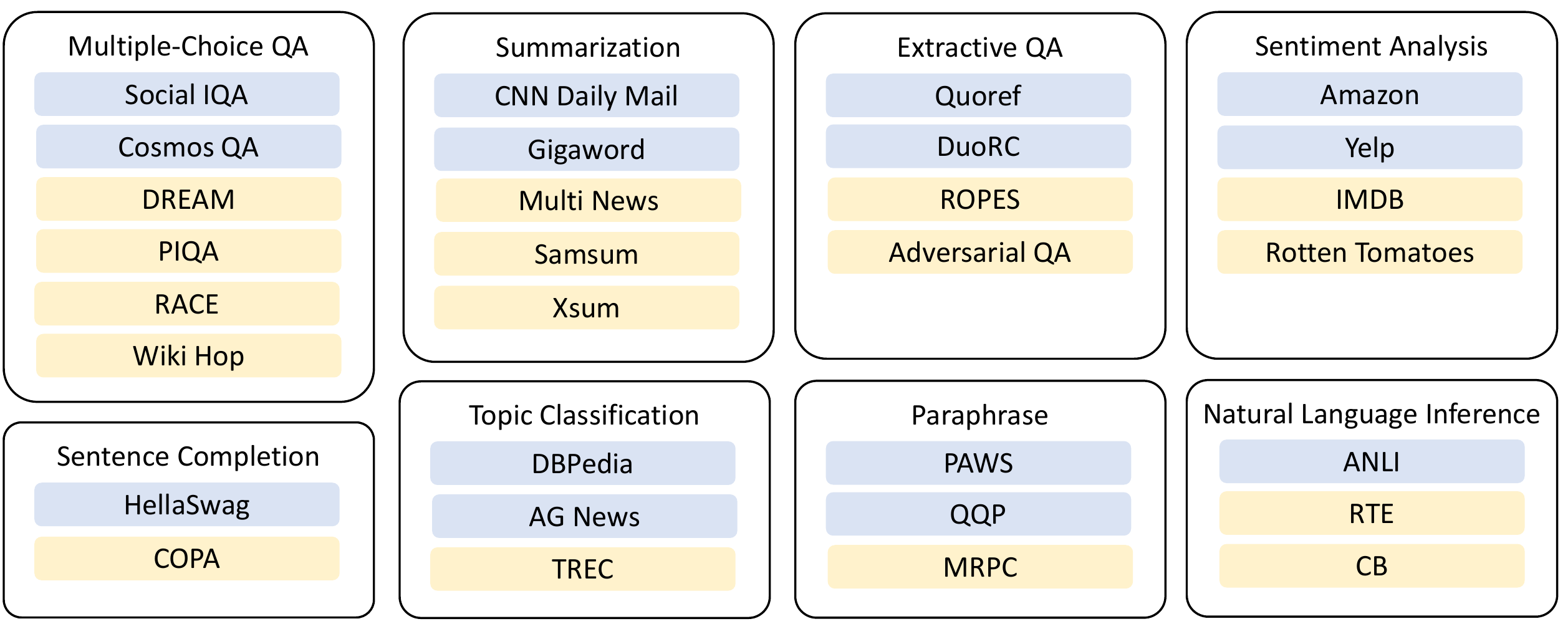}
		\caption{The task taxonomy of the tasks (datasets). The tasks used in the multi-task pre-training mixture are marked in blue.  Yellow tasks are unseen during pre-training and are used for downstream evaluation under both zero-shot testing and few-shot learning settings.}
% 			We omit the fused evidence block from the figure for simplification.}
		\label{fig:tasksplit}
	\end{figure*}

In this part, we introduce the task taxonomy, and the underlying scenario for measuring the task generalization ability of the schema prompt.
% We begin by clearly introduce some definitions of the task generalization. 
% In our work, we define ``task'' as a specific dataset, and ``format'' as a common task format of a group of tasks. 
% For example, ``\textit{DREAM}'' is defined as a task, and its corresponding format is ``\textit{Multiple Choice QA}''. 
% Noting that the underlying reason is that different datasets always emphasize different kinds of required skills, and the data instances in various datasets have very diverse distribution. 
% Thus, we largely follow \citet{brown2020language} and adopt the similar definition. 
To test the task generalization ability on various NLP tasks at scale, we use 30 publicly available benchmark NLP tasks, belonging to 8 formats (e.g., \textit{QA, NLI, Summarization, etc.}) for our experiment.
As the task taxonomy shown in Fig. \ref{fig:tasksplit}, we select several tasks (marked in blue) for multi-task prompted pre-training, and take the rest tasks (marked in yellow) that are unseen during pre-training for downstream evaluation.  
%Note that the unseen evaluation task mainly belongs to previously learned formats in our task taxonomy. 
\paragraph{Rationale for Task Taxonomy.}
The underlying reasons for using this taxonomy criterion are listed as follows.
(1) In most real-world applications, model adaptation to an unseen task (with new data distribution or domain knowledge) is a much more frequent practice than adapting to a completely new format type (like QA or NLI). 
(2) Since Schema Prompt-based pre-training helps the model in learning prior knowledge about input components and task attributes, pre-training and evaluation on the similar format type is the best way to utilize the learned knowledge. It worth noting that we adopt this setup (generalization to unseen tasks with seen formats) for our main experiment (\cref{sec:main-experiment}). We further conduct comprehensive experiments to explore the generalization ability of \modelname~to unseen format (unseen tasks with unseen format), as detailed in \cref{sec:task-composition} and Appendix A. 
% (1) Since both the semantic meaning of the key indicators and format-specific component are learnable in the schema prompt, learning the prior knowledge about them in the pre-training stage will benefit following task generalization.
% (2) In most real-world applications, task adaptation to a new task with unseen data distribution or domain knowledge is more commonly required than adapting to a completely unseen format type.

% adaptation to the unseen data distribution rather than a new format type is can better reflect the real-world applications.
% More experiments on adaptation to unseen format types are given in Appendix A.

The whole paradigm of adopting schema prompt for task generalization consists of following procedures.
We first reformulate the inputs and outputs for each task using the unified schema prompt, and construct the mixed schema-based prompted pre-training corpus. 
% For multi-task prompted pre-training, we adopt the reformulated instances in all tasks from the training mixture, to train a unified encoder-decoder model together with the schema prompt. 
After pre-training corpus construction, we pre-train a unified encoder-decoder model together with learnable parameters residing in the schema prompt. 
At this end, both the commonly shared knowledge across tasks and the semantic meaning of schema prompt are learned as a prior. 
% For multi-task prompted pre-training, we train a unified encoder-decoder model together with mixed structurally-reformatted instances from all tasks.
For evaluating the task generalization ability, we measure the effectiveness of the schema prompt on tasks unseen during pre-training, under both the zero-shot testing and few-shot learning settings. 
The zero-shot testing setting evaluates the zero-shot generalization to unseen task while the few-shot learning setting aims to measure the effectiveness and performance of low-resource model adaption to a newly involved task.
% We conduct evaluation under both the zero-shot testing and few-shot learning settings. 
% The former targets to evaluate the zero-shot generalization to unseen data distribution, and the latter aims to measure the effectiveness and performance of low-resource model adaption to a newly involved task.

\section{Experiments}
\subsection{Experimental Setup}
\paragraph{Model Architecture}
% The multi-task learning framework is built based on the encoder-decoder architecture. 
% Specifically, we feed the reformatted inputs in the multi-task training mixture as the inputs of the encoder, and ask the decoder to generate corresponding outputs. 
We set T5 \citep{raffel2020exploring} as the backbone of the encoder-decoder model. 
T5 is a strong Transformer-based language model that is pre-trained with C4. 
We adopt \texttt{google/t5\--v1\_1\--base} from HuggingFace Transformers \citep{wolf-etal-2020-transformers} which is only pre-trained on C4 excluding any supervised data.

\paragraph{Training}
Our model, namely \modelname, is trained with the training mixture as described in Section~\ref{sec:task_generalization}. 
% We train the model with 10 epochs and don't perform checkpoint selection and selects the last checkpoint for downstream evaluation. 
To balance the number of instances in datasets, we set a constraint to each dataset in which the maximum number of training instances should be smaller than 700,000.
% set the maximum number of training instances as 700,000. 
It is worth noting that the learnable parameters in the schema prompt are learned together with the parameters of T5. 
The groups of soft key prompts are different and independent for different component types.
Similarly, the format/task/output-specific values (also learnable soft prompts) are also independent for different format/task/output types.
The detailed hyper-parameters and the dimension of the soft key prompts and special task-attributed value prompts are given in Appendix B.

\paragraph{Evaluation}
We evaluate the task generalization ability on the unseen evaluation tasks (i.e., 16 yellow tasks in Fig.~\ref{fig:tasksplit}) under both zero-shot testing and few-shot learning settings. 
During few-shot learning, we adopt standard few-shot learning strategy that utilizes 32 randomly selected instances from each task for low-resource model adaptation.
We evaluate the performance on the validation set of each task, or the test set if the validation set is not available. 
% Specifically, we introduce how to initialize the schema prompt-formulated input for evaluation. 
For the soft prompts corresponding to each common format/output type and soft key prompts that are seen during training, we directly initialize these learned prompts
% for the specific format/output-specific values and key indicators 
for evaluation. 
Since the task is unseen during training, the \texttt{value} (prompts) under the \texttt{[Task]} key will be randomly initialized.
% for zero-shot testing and the initialization of few-shot learning.  

\paragraph{Metrics}
% Here we introduce the evaluation metric for different tasks. 
For the task belonging to ``\textit{Extractive QA}'' that requires to extract an answer from the passage, we adopt commonly used \textit{exact match (EM)} as the evaluation metric. 
For the tasks required to generate a free-formed description from the given context (e.g., ``\textit{Summarization}''), we adopt \textit{Rouge-L} as the evaluation metric. 
For the rest tasks that involve choosing the best answer from several given candidate options (e.g., ``\textit{Multiple Choice QA}'', ``\textit{Topic Classification}'', etc.), we adopt \textit{accuracy} as the metric. 
To calculate the scores of options for the classification tasks, we follow \citet{sanh2021multitask} and take the log-likelihood of each option as the score for options ranking, and select the option with the highest log-likelihood as the final answer.

\paragraph{Baselines}
In this work, we mainly target at comparing the task generalization ability between the NL prompt and schema prompt in the  multi-task learning paradigm. 
Therefore, we adopt the reliable NL prompts source collected by T0 \citep{sanh2021multitask}, which introduces the most relevant and powerful NL prompted-based multi-task pre-training method.
T0 adopts a crowd-sourcing platform to collect human-written NL prompts as templates to reformulate inputs and outputs of different tasks, and performs multi-task prompted pre-training. 
It collects a diverse set of NL prompts for each task, and the resulted collection is noted as \textit{Public Pool of Prompts (P3)}, which is mostly publicly available\footnote{\url{https://huggingface.co/datasets/bigscience/P3}}. 
Therefore, we pretrain two NL prompt-based baselines (\textit{NLPro-single} and \textit{NLPro-multi}) using \textit{P3} for us to directly compare the effectiveness between the NL prompt and the schema prompt in task generalization.
Note that \textit{NLPro} is different from T0 in the task taxonomy and model size.
% with the same hyper-parameters and same supervisions as our baselines for fair comparison:
% believe that it is the most suitable baseline for us to directly compare the effectiveness between the NL prompt and the schema prompt in task generalization.
% Since the task taxonomy and model size of T0 is different from our work, we pre-train the two variants of T0 (\textit{T0-single} and \textit{T0-multi}) 
We use the same hyper-parameters and same supervisions as our method for fair comparison.
% To be consistent with T0, the NL prompt-formatted instances of each task are collected from the released \textit{P3} dataset.
(1) \textit{NLPro-single}: In this variant, we adopt a single NL-prompt from \textit{P3} to reformulate each task completely. We randomly select the prompt from the collections for each task. 

(2) \textit{NLPro-multi}: To increase the diversity of NL prompts and improve the consistency with the original settings in T0, we also adopt multiple prompts (denoted as \texttt{prompt\_number}) for each task in the pre-training mixture. 
During pre-training, we split each training dataset randomly into \texttt{prompt\_number} parts\footnote{We don't repeat the whole dataset  for each prompt in training because we want to avoid the bias of data augmentation for fair comparison.} and formulate each part with the corresponding prompt.  We set maximum \texttt{prompt\_number} as 3 per task for training. 
During Evaluation, we report the averaged scores of all single tested prompts.

\subsection{Main Results}
\label{sec:main-experiment}
% We aims to explore whether schema prompt can enhance task generalization ability on the unseen tasks with the multi-task learning paradigm. 
% We report the main results on Table \ref{tab:main-result}. 
% Our method is abbreviated as \modelname. We compare different variants of T0 (\textit{T0-single} and \textit{T0-multi}) with the our method, and report results on 16 evaluation tasks, belongs to 8 format types (e.g., \textit{Multiple Choice QA, Summarization, Extractive QA, NLI, etc.}). 
% We test the task generalization ability under both zero-shot testing and few-shot learning settings.
% % Table generated by Excel2LaTeX from sheet 'Overall'
\begin{table}[t!]
\caption{Main results on 16 evaluation tasks belonging to 8 formats, under both zero-shot testing and few-shot learning settings. }
\label{tab:main-result}
\vspace{0.1in}
  \centering
     \resizebox{1.0\textwidth}{!}{
\begin{tabular}{ccc|ccc|ccc}
    \Xhline{3\arrayrulewidth}
    \multirow{2}[2]{*}{Task} & \multirow{2}[2]{*}{Metric} & \multirow{2}[2]{*}{Dataset} & \multicolumn{3}{c|}{Zero-shot} & \multicolumn{3}{c}{Few-shot} \bigstrut[t]\\
         &      &      & NLPro-single & NLPro-multi & \modelname & NLPro-single & NLPro-multi & \modelname \bigstrut[b]\\
    \hline
    \multirow{4}[2]{*}{MultiQA} & \multirow{4}[2]{*}{Acc.} & DREAM & 47.16 & 43.14 & \textbf{58.24} & 54.75 & 50.27 & \textbf{59.65} \bigstrut[t]\\
         &      & PIQA & 49.62 & 49.62 & \textbf{58.32} & 54.95 & 56.01 & \textbf{58.71} \\
         &      & RACE & 31.96 & 37.44 & \textbf{42.05} & 35.70 & 37.92 & \textbf{42.19} \\
         &      & WikiHop & 14.37 & 14.92 & \textbf{16.37} & 17.17 & 14.08 & \textbf{30.27} \bigstrut[b]\\
    \hline
    \multirow{2}[2]{*}{Extractive QA} & \multirow{2}[2]{*}{EM} & ROPES & 30.45 & 28.85 & \textbf{37.32} & 47.09 & 37.56 & \textbf{50.59} \bigstrut[t]\\
         &      & Adversarial QA & 20.40 & 18.80 & \textbf{24.50} & 22.70 & 22.90 & \textbf{27.20} \bigstrut[b]\\
    \hline
    \multirow{2}[2]{*}{Sentiment } & \multirow{2}[2]{*}{Acc.} & IMDB & 92.90 & 93.55 & \textbf{95.05} & 93.46 & 93.20 & \textbf{95.89} \bigstrut[t]\\
         &      & Rotten Tomatoes & 57.97 & 69.80 & \textbf{89.68} & 86.49 & 86.43 & \textbf{90.81} \bigstrut[b]\\
    \hline
    Topic Class. & Acc. & TREC & \textbf{27.60} & 18.93 & 24.60 & 72.00   & 62.67 & \textbf{76.20} \bigstrut\\
    \hline
    Paraphrase & Acc. & MPRC & 31.62 & 37.42 & \textbf{72.30} & 68.63 & 68.37 & \textbf{75.49} \bigstrut\\
    \hline
    \multirow{3}[2]{*}{Summarization} & \multirow{3}[2]{*}{RougeL} & Multi News & 6.42 & 5.88 & \textbf{6.16} & \textbf{6.62} & 6.22 & 6.53 \bigstrut[t]\\
         &      & Samsum & 10.70 & 10.15 & \textbf{20.32} & 30.39 & 30.07 & \textbf{32.85} \\
         &      & Xsum & 11.81 & 10.41 & \textbf{12.86} & 15.28 & 18.42 & \textbf{18.94} \bigstrut[b]\\
    \hline
    Sen. Comp. & Acc. & COPA & 61.00   & 61.60 & \textbf{62.00} & 66.00   & 65.00   & \textbf{72.00} \bigstrut\\
    \hline
    \multirow{2}[2]{*}{NLI} & \multirow{2}[2]{*}{Acc.} & RTE  & 75.81 & 72.68 & \textbf{80.87} & 76.80 & 73.85 & \textbf{83.03} \bigstrut[t]\\
         &      & CB   & 83.93 & 68.75 & \textbf{85.71} & 85.71 & 82.14 & \textbf{91.07} \bigstrut[b]\\
    \hline
    Average &   -   &   -   & 40.86 & 40.12 & \textbf{49.15} & 52.11 & 50.32 & \textbf{56.96} \bigstrut\\
    \Xhline{3\arrayrulewidth}
\end{tabular}
    }
\vspace{0.1in}

\end{table}%

Zero-shot testing and few-shot learning results on 16 unseen tasks from 8 formats are shown in Table~\ref{tab:main-result} \footnote{To further explore whether schema prompt is beneficial for unseen formats, we also experiment with task taxonomy that training and evaluation are separately conducted on different formats, and report results on Appendix B.}.
% To verify whether schema prompt is still beneficial when there is enough supervised data, we also experiment under the full-data fine-tuning setting, and report results on Appendix B.}. 
Our observations are listed as follows:
\begin{itemize}[leftmargin = 15pt,topsep=0pt,noitemsep]
    \item Our approach \modelname\ outperforms NL prompt-based methods on 15 out of 16 tasks. 
On average, \modelname\ significantly improves the zero-shot testing and few-shot learning performance by 8.29\% and 4.85\% respectively, demonstrating better task generalization capability than NL prompt.
    \item \modelname\ enables better modeling the transferable knowledge across different tasks because it helps the model to explicitly identify the components with learnable key indicators and thus can learn the general semantics of component types.
    \item The format/task-specific values customize the knowledge specialized for each format type and task, which is essential in helping the model to restore the knowledge required for each task, and better discriminating them. 
    \item \textit{NLPro-single} and \textit{NLPro-multi} results exhibit large performance variance using different NL prompts in many tasks, which indicates that various NL prompts may lead to instability when adapting to unseen tasks \citet{sanh2021multitask}.
    % , which is consistent with the findings in \citet{sanh2021multitask}.
\end{itemize}

% on the average of 16 tasks, and improves the few-shot learning performance by 4.85\% points on average.
% This observation demonstrates better task generalization than T0. 
% More importantly, the experiment results also show that schema prompt performs better than NL-based prompts because of the following reasons. 
% Firstly, \modelname\ enable better modeling the transferable knowledge across different tasks, because it helps the model to explicitly identify the components with learnable key indicators and thus can learn the general semantics of component types.
% Secondly, the format/task-specific values customize the knowledge specialized for each format type and task, which is essential in helping the model to restore the knowledge required for each task, and better discriminating them. 
% Moreover, the results also exhibit large performance variance between \textit{T0-single} and \textit{T0-multi} in many tasks, which indicates that various NL prompts may lead to instability when adapting to unseen tasks, which is also consistent with the findings in \citet{sanh2021multitask}.

\subsection{Ablation Study}

To evaluate the effectiveness of involving learnable key indicators and task-attribute specific components (i.e., \textit{Task, Format}) as learnable \textit{key-value} pairs into the schema prompt, we conduct three ablation experiments: 
(1) removing the format-specific \textit{key-value} (\modelname\ w/o F);
(2) removing the task-specific \textit{key-value} (\modelname\ w/o T); 
(3) removing the learnable key indicator (\modelname\ w/o K).
We report results on 8 unseen tasks under both zero-shot and few-shot settings in Table~\ref{tab:ablation}.

\paragraph{Effect of Format-specific Prompt}
% The results of removing format-specific prompts are shown in Table \ref{tab:ablation}.
Removing format-specific prompt leads to significant performance drop, showing that format-specific prompt enables learning format-specific knowledge during multi-task pre-training and provides guidance to the downstream tasks. 
% (Note: Zero-shot: format is much more essential than task)

\paragraph{Effect of Task-specific Prompt}
Removing task-specific performance prompt largely harms the performance on all tasks under all settings, especially in the few-shot learning settings.
This observation verifies that it is important to record the specialized knowledge for each task, as different tasks require different kinds of knowledge (e.g., ``\textit{commonsense reasoning}'') or have different data distribution. 
Noting that the effect of format prompt is more significant than task prompt in the zero-shot setting, because the format-specific knowledge is already learned during pre-training and task knowledge is unknown for the unseen task without few-shot training. 
However, the task prompt is helpful in discriminating different tasks (even a new task), which is beneficial for zero-shot task generalization.

% (Note: Few-shot: both format and task are all very essential)
\paragraph{Effect of Learnable Key Prompts}
Removing the special key prompts from the schema prompt harms the model in terms of identifying the different input components of each task. Therefore, it performs worse because it is harder to model the common knowledge of tasks and discriminate different components.
 % Table generated by Excel2LaTeX from sheet 'Overall'
% Table generated by Excel2LaTeX from sheet 'Overall'
% Table generated by Excel2LaTeX from sheet 'Overall'
% Table generated by Excel2LaTeX from sheet 'Overall'
\begin{table}[h]
\centering
  \caption{Ablation study under zero-shot testing and few-shot learning settings on 8 datasets. ``w/o F/T" indicates eliminating the format/task components from the schema prompt. ``w/o K" indicates eliminating learnable key prompts.}
  \label{tab:ablation}
  \vspace{0.1in}
\resizebox{1.0\textwidth}{!}{
    \begin{tabular}{c|l|cc|cc|c|c|c|c|c}
    \Xhline{3\arrayrulewidth}
    \multirow{2}[4]{*}{Setting} & \multicolumn{1}{c|}{\multirow{2}[4]{*}{Model}} & \multicolumn{2}{c|}{MultiChoiceQA} & \multicolumn{2}{c|}{Extractive QA} & Sentiment & \multicolumn{1}{l|}{Para.} & Summary & NLI  & \multirow{2}[4]{*}{Avg.} \bigstrut\\
\cline{3-10}         &      & DREAM & WikiHop & ROPES & Adv. QA & IMDB & MRPC & Samsum & RTE  &  \bigstrut\\
    \hline
    \multirow{4}[2]{*}{Zero-shot} & \modelname & 58.2 & 16.4 & 37.3 & 24.5 & 95.1 & 72.3 & 20.3 & 80.9 & 50.6 \bigstrut[t]\\
         & - w/o F & 55.3 & 14.4 & 31.4 & 22.8 & 92.8 & 68.9 & 16.0 & 75.5 & 47.1 \\
         & - w/o T & 56.6 & 15.0 & 32.9 & 23.1 & 93.8 & 70.8 & 17.7 & 76.5 & 48.3 \\
         & - w/o K & 56.8 & 15.2 & 30.2 & 20.2 & 93.7 & \multicolumn{1}{c|}{70.6} & \multicolumn{1}{c|}{19.0} & \multicolumn{1}{c|}{75.3} & \multicolumn{1}{c}{47.6} \bigstrut[b]\\
       
    \hline
    \multirow{3}[2]{*}{Few-shot} & \modelname & 59.7 & 30.3 & 50.6 & 27.2 & 95.9 & 75.5 & 32.9 & 83.0 & 56.9 \bigstrut[t]\\
         & - w/o F & 58.4 & 29.4 & 49.2 & 25.7 & 94.9 & 71.6 & 32.0 & 78.3 & 54.9 \\
         & - w/o T & 57.2 & 27.6 & 44.8 & 25.3 & 94.9 & 72.8 & 32.1 & 79.6 & 54.3 \\
         & - w/o K & 57.4 & 26.6 & 43.5 & 25.2 & 94.8 & 71.5     &  32.1   &   79.0   &  53.8 \bigstrut[b]\\
    \Xhline{3\arrayrulewidth}
    \end{tabular}
    }
    \vspace{0.1in}
  
\end{table}%

% \begin{table}[htbp]
% \small
%   \centering
%     \begin{tabular}{c|l|cc|cc|c|c|c|c|c}
%     \toprule
%     \multirow{2}[4]{*}{Setting} & \multicolumn{1}{c|}{\multirow{2}[4]{*}{Model}} & \multicolumn{2}{c|}{MultiChoice QA} & \multicolumn{2}{c|}{Extractive QA} & \multicolumn{1}{c}{Sentiment} & \multicolumn{1}{l|}{Para.} & Summary & NLI   & \multirow{2}[4]{*}{Avg} \\
% \cmidrule{3-10}          &       & DREAM & WikiHop & ROPES & Adv. QA & \multicolumn{1}{c}{IMDB} & MRPC  & Samsum & RTE   &  \\
%     \midrule
%     \multirow{3}[2]{*}{Zero-shot} & \modelname & 58.2  & 16.4  & 37.3  & 24.5  & 95.1  & 72.3  & 20.3  & 80.9  & 50.6 \\
%           & \modelname\ w/o F & 55.3  & 14.4  & 31.4  & 22.8  & 92.8  & 68.9  & 16.0  & 75.5  & 47.1 \\
%           & \modelname\ w/o T & 56.6  & 15.0  & 32.9  & 23.1  & 93.8  & 70.8  & 17.7  & 76.5  & 48.3 \\
%     \midrule
%     \multirow{3}[2]{*}{Few-shot} & \modelname & 59.7  & 30.3  & 50.6  & 27.2  & 95.9  & 75.5  & 32.9  & 83.0  & 56.9 \\
%           & \modelname\ w/o F & 58.4  & 29.4  & 49.2  & 25.7  & 94.9  & 71.6  & 32.0  & 78.3  & 54.9 \\
%           & \modelname\ w/o T & 57.2  & 27.6  & 44.8  & 25.3  & 94.9  & 72.8  & 32.1  & 79.6  & 54.3 \\
%     \bottomrule
%     \end{tabular}
%   \label{tab:addlabel}
%   \caption{Ablation Study under zero-shot testing and few-shot learning settings on 8 datasets. ``w/o F/T" indicates the eliminating the format/task-specific key-value pairs from the structural prompt.}
% \end{table}

\subsection{Task Compositionality with Key Prompts}
\label{sec:task-composition}
In this part, we target on exploring whether identifying the semantic meaning of different components (keys) can actually benefit task generalization. 
% We begin by defining a scenario \textit{task compositionality} to investigate this problem. 
We design a scenario to investigate the effect of key prompts:
\textit{''Once \modelname\ learns two formats $A$ and $B$ with components types $K_A$ and $K_B$, can it generalize the learned semantic meaning of components $K_A$ and $K_B$ to an unseen format $C$ with compositional components $K_C=K_A\cup K_B$, with only a few examples?''}

To answer this question, we set a specific scenario of \textit{task compositionality}: we utilize tasks belonging to two formats $A$ and $B$ for model training, and evaluate on an unseen format with compositional components from these two formats. 
Specifically, we train our model with the combinations of 3 tasks (i.e., \textit{QuoRef, DuoRC and ROPES}) belonging to format $A=\textit{Extractive QA}$ with components $K_{A}=\{\textit{passage, question}\}$, and 3 tasks \textit{(i.e., AgNews, DBPedia and IMDB)} belonging to format $B=\textit{Text Classification}$ with components $K_{B}=\{\textit{passage, options}\}$. 
% Then we evaluate on the new format $C=\textit{Multiple Choice QA}$, with compositional components $K_C=K_A\cup K_B=\{\textit{passage,question,options}\}$.
During evaluation, we adopt 6 tasks (i.e., \textit{DREAM, PIQA, RACE, WikiHop, Cosmos QA and Social IQA}) belonging to new format $C=\textit{Multiple Choice QA}$, with compositional components $K_C=K_A\cup K_B=\{\textit{passage, question, options}\}$.

% Table generated by Excel2LaTeX from sheet 'Overall'
% Table generated by Excel2LaTeX from sheet 'Overall'
\begin{table}[t]
\caption{Task compositionality experiment. 
  The model is trained on the combination of 3 datasets \textit{(QuoRef, DuoRC, ROPES)} from extractive QA task with components ``\textit{\{passage, question\}}", and 3 classification datasets \textit{(AgNews, DBPedia, IMDB)} with components ``\textit{``\{passage, options\}"}". 
  The model is evaluated on 6 multiple choice QA datasets with compositional components ``\textit{\{passage, question, options\}}" from the learned tasks, to explore the compositionality of tasks.}
  \label{tab:compositionality}
  \vspace{0.1in}
  \centering
\resizebox{\textwidth}{!}{    \begin{tabular}{cc|cccccc}
    \Xhline{3\arrayrulewidth}
    Setting & Prompt Type & \multicolumn{6}{c}{Dataset} \bigstrut\\
    \hline
         &      & DREAM & PIQA & RACE & WikiHop & Cosmos QA & Social IQA \bigstrut\\
    \hline
    \multirow{2}[2]{*}{Zero-shot} & NL Prompt & 34.2 & 51.9 & 22.1 & 12.9 & 25.1 & 33.9 \bigstrut[t]\\
         & \modelname & 35.1 & 49.5 & 27.1 & 11.8 & 28.0 & 34.6 \bigstrut[b]\\
    \hline
    \multirow{2}[2]{*}{Few-shot} & NL Prompt & 35.8 & 50.3 & 26.6 & 13.2 & 30.8 & 37.6 \bigstrut[t]\\
         & \modelname & 39.3 & 51.4 & 30.9 & 25.0 & 38.4 & 41.0 \bigstrut[b]\\
    \Xhline{3\arrayrulewidth}
    \end{tabular}}
    \vspace{0.1in}

\end{table}

% We compare with baseline that using only NL prompt for model training and generalization (the same setting with \textit{T0-single}), to compare the task generalization ability of NL prompt and schema prompt on compositional tasks. 
We compare \modelname\ with NL prompt on the compositional scenario. 
As the results shown in Table \ref{tab:compositionality}, schema prompt achieves better performance than NL prompt in 4 of 6 held-out compositional tasks under the zero-shot testing setting, and significantly outperforms NL prompt in all tasks under few-shot learning setting.
This supports our hypothesis that learning semantics of components in an explicit way can benefit task generalization, even for an unseen format type.
Since our method has already explicitly learned the semantics of components $K_A$ and $K_B$, we can teach the model about their compositional semantics with only a few examples, to make faster and better generalization. 
Noting that the relatively weak zero-shot performance is reasonable, because NL prompt can provide additional human-written instruction to tell the model how to solve a task belonging to a completely unseen format type with unknown reasoning skills. 
% , which is lack in the schema prompt. 
\subsection{Full-data Fine-tuning}
The aforementioned experiments demonstrate better task generalization ability of schema prompt in the low resource settings.
% zero-shot and few-shot learning settings. 
We are still curious about whether the schema prompt is beneficial when there is enough supervised training data for downstream tasks? 
% We aims to explore whether the schema prompt is still beneficial when there is enough supervised training data for downstream tasks? 
To answer this question, we also conduct experiments under the full-data fine-tuning setting on 7 downstream tasks that are unseen during multi-task pre-training, and report results in Table \ref{tab:fulldata}.
As shown in the table, schema prompt demonstrates better performance than NL prompt (\textit{NLPro-single}) on these 7 downstream evaluation tasks.
This observation shows that the shared knowledge and the discriminating ability of different components and task attributes modeled by the schema prompt are still essential for model learning, even there is enough supervised data. 
This finding also indicates broadening potential applications of the schema prompt as a unified input schema. 
% showing that explicitly modeling different components and the task attributes is essential under the full-data fine-tuning setting.
% Table generated by Excel2LaTeX from sheet 'Overall'
\begin{table}[htbp]
  \centering
  \caption{Results on 7 downstream tasks under the full-data fine-tuning setting. }
  \label{tab:fulldata}
  \vspace{0.1in}
  \resizebox{1.0\textwidth}{!}{
    \begin{tabular}{cc|ccccccc}
    \Xhline{3\arrayrulewidth}
    Setting & Prompt Type & \multicolumn{6}{c}{Task}                &  \bigstrut\\
    \hline
         &      & DREAM & RACE & ROPES & Adversarial QA & Rotten Tomatoes & Samsum & COPA \bigstrut\\
    \hline
    \multirow{2}[1]{*}{Full-Data} & NL Prompt & 69.4 & 61.2 & 53.8 & 33.7 & 88.5 & 40.2 & 71.0 \bigstrut[t]\\
         & \modelname & 72.4 & 70.1 & 54.8 & 35.4 & 90.7 & 41.0   & 73.0 \\
    \Xhline{3\arrayrulewidth}
    \end{tabular}
    }
    \vspace{0.1in}
  
\end{table}%

\section{Related Work}

Prompt-learning~\citep{liu2021pre} on pre-trained language models~\citep{devlin2018bert, raffel2019exploring, brown2020language, Han2021PreTrainedMP} has demonstrated effectiveness on a wide range of NLP tasks under the few-shot and zero-shot settings. 
The primal prompting adapted by GPT-3~\citep{brown2020language} does not involve parameter updates, but simply introduces additional contexts to perform "in-context learning" to obtain promising results in low data scenarios. 
Then a subsequent series of methods show that projecting downstream tasks to pre-training tasks via manually written or automatically generated prompts is effective on pre-trained language models across different sizes and structures~\citep{shin-etal-2020-autoprompt, schick-schutze-2021-exploiting, schick-schutze-2021-just,gao-etal-2021-making, le2021many, ding2021openprompt}, especially when labeled data is insufficient. 
Prompts are not necessarily textual, some works develop prompts in continuous space~\citep{li-liang-2021-prefix, zhong2021useradapter,lester2021power, liu2021gpt, qin2021exploring, liu2021p} and it is found that such soft prompts could not only represent vague semantics, but also serve as a parameter-efficient method~\citep{he2021towards, ding2022delta} to fine-tune pre-trained language models.
In addition to evaluation on separate NLP tasks, prompting is also explored in multi-task scenarios~\citep{sanh2021multitask, xu2022zeroprompt}. 
ProQA \citep{https://doi.org/10.48550/arxiv.2205.04040} adopts structurally-designed prompt to unify QA tasks. However, it targets at using minimal supervision to build a general QA model and only focuses on QA tasks, while our work focuses on improving task generalization ability for general NLP tasks at scale and involves more complicated task schemas. 
T0~\citep{sanh2021multitask} trains a sequence-to-sequence model with a number of human-written prompts guided by professional crowd-source instructions and shows that such a model could show remarkable capability of zero-shot generalization on held-out NLP tasks.
Our work also explores low-resource prompt-based task generalization, but based on an automatically constructing strategy according to the data schemas. In terms of the constructing process, our approach relies only on some explicit information of datasets, thus eliminating a large amount of overhead in writing diverse prompts. 
It is also worth noting that although the cost of writing a prompt is greatly reduced, our schema differs from T0's due to the fact that this automatically constructed schema prompt requires knowledge transferring across tasks under a broad category.
\section{Discussion}
We target on discussing the limitations of our approach, and exploring the potential future direction of \modelname\ to shed a light on future directions.
  
As we mentioned before, \modelname\ is capable of modeling common knowledge shared across tasks by learning explicit prior knowledge about shared schema and task attributes (i.e., \textit{format} and \textit{output}), and enhancing task generalization ability. 
Intuitively, our model will be weakened in task generalization to a completely new format type with no ever presented component types. 
In this case, natural language prompts can provide some human-written guidance to hint the model in problem solving.

Furthermore, we point out some interesting future directions for extending \modelname.
Firstly, \modelname\ can be adopted to more modalities. Multi-modal tasks can involve complex input schema with components from different modalities, e.g., \textit{video, language, image, audio, etc.} \modelname\ can be utilized to flexibly compose inputs from different modalities and discriminate their variances.
Secondly, \modelname\ can be extended to store supported knowledge as a new component. Solving many realistic problems requires to retrieve and use knowledge from different domains \cite{zhongreasoning, zhong2019reasoning,hu-etal-2021-compare} (e.g., \textit{tables, passages, knowledge graphs}). Knowledge type and retrieved knowledge can also be extended as learnable components, to share knowledge across domains and also discriminate them.
Moreover, \modelname\ can have hierarchical structure. That means, the value in it can have nested components to store fine-grained information. For example, we can parse \textit{POS tags} or \textit{Entity type} for a textual value, and take them as nested components under this value, to give fine-grained clues to the model.

\section{Conclusion}
This paper improves task generalization ability of NLP tasks, with a unified schema-based prompting method - \modelname, which is capable of automatically constructing the prompt according to the task schema, modeling the shared knowledge across tasks, and simultaneously capturing their specialties. 
Our approach \modelname\ conducts schema prompt-based multitask pre-training and achieves strong zero-shot and few-shot performance on 16 unseen downstream tasks.
Further analyses demonstrate the effectiveness of each component residing in the schema prompt, shows that it is more flexible in model adaptation to compositional tasks, and has better performance in the full-data setting. 

\bibliography{anthology,custom}
\bibliographystyle{nips}
%%%%%%%%%%%%%%%%%%%%%%%%%%%%%%%%%%%%%%%%%%%%%%%%%%%%%%%%%%%%
% \input{checklist}
%%%%%%%%%%%%%%%%%%%%%%%%%%%%%%%%%%%%%%%%%%%%%%%%%%%%%%%%%%%%
\appendix
\section{Out-of-Format Analysis}
\begin{figure*}[h]
		\centering
		\includegraphics[width=\textwidth]{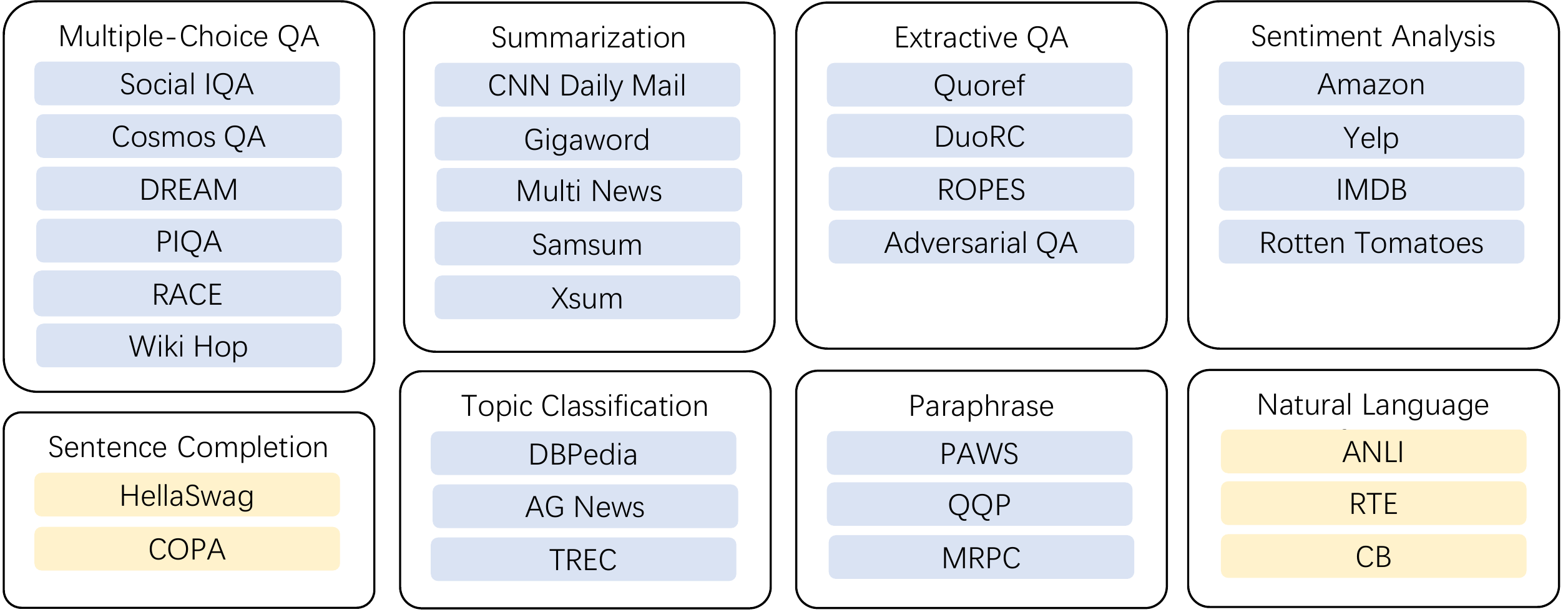}
		\caption{The task taxonomy of the tasks (datasets) for the out-of-format analysis. The tasks used in the multi-task pre-training mixture are marked in blue.  Yellow tasks are unseen during pre-training, and are used for downstream evaluation under zero-shot setting.}
% 			We omit the fused evidence block from the figure for simplification.}
		\label{fig:task-split-oof}
	\end{figure*}
In this part, we target to investigate the effectiveness of the schema prompt during zero-shot testing on a completely unseen format type. 
The task taxonomy of the out-of-format analysis is shown in Fig. \ref{fig:task-split-oof}.
We select tasks belongs to ``\textit{Sentence Completion}" and ``\textit{Natural Language Inference}" for evaluation, and the rest tasks are used for multi-task pre-training. 
We compare with \textit{NLPro-multi}, and report the averaged performance (NLPro-multi (AVG)), and the standard deviation of all the results (NLPro-multi (STD)) of all the tested NL prompts.
The results are reported in Table \ref{tab:oof-analysis}. 
Noting that to better enhance prior knowledge learned from \modelname, we use component types \textit{\{sentence1, sentence2\}} to represent the task schema of \textit{Sentence Completion, NLI and Paraphrase}.
It can be observed that \modelname\ still outperforms NLPro-multi in most of the tasks, excluding \textit{RTE}, showing that \modelname\ is still more effective in modeling the common knowledge shared across formats. 
Moreover, it can be also observed that the standard deviation of results on the all tested NL prompts in NLPro-multi is high for many tasks (e.g., 13.47\% for \textit{CB} task, and 8.99\% for \textit{COPA} task). This finding also shows that different NL prompts may lead to larger performance variance on downstream tasks.

% Table generated by Excel2LaTeX from sheet 'Overall'
\begin{table}[htbp]
  \centering
    \begin{tabular}{c|ccccc}
    \Xhline{3\arrayrulewidth}
    Method & \multicolumn{5}{c}{Task} \bigstrut\\
    \hline
         & Hellaswag & RTE  & CB   & COPA & ANLI \bigstrut\\
    \hline
    \modelname & 31.14 & 58.6 & 62.5 & 63   & 33.6 \bigstrut\\
    \hline
    NLPro-multi (AVG) & 27.28 & 62.5 & 30.95 & 61.6 & 31.71 \bigstrut[t]\\
    NLPro-multi (STD) & 0.99 & 6.69 & 13.47 & 8.99 & 1.2 \bigstrut[b]\\
    \Xhline{3\arrayrulewidth}
    \end{tabular}
\vspace{0.1in}  \caption{The results of zero-shot testing on tasks belongs to unseen formats. NLPro-multi (AVG) is the averaged performance, and NLPro-multi (STD) is the standard deviation of all the tested NL prompts.}
  \label{tab:oof-analysis}
\end{table}%

\iffalse
\section{Full-data Fine-tuning}
We aims to explore whether the schema prompt is still beneficial when there is enough supervised training data for downstream tasks? 
We conduct experiments under the full-data fine-tuning settings on 7 downstream tasks that are unseen during multi-task pre-training, and report results on Table \ref{tab:fulldata}.
schema prompt demonstrates better performance than NL prompt (NLPro-single) on these 7 downstream evaluation tasks, showing that explicitly modeling different components and the task attributes is essential under the full-data fine-tuning setting.
\input{tables/fulldata}
\fi
\section{Implementation Details}
\subsection{Multi-task Pre-training}
During multi-task pre-training, we formulate each task with the schema prompt, and construct the multi-task pre-training corpus.
We map each type of key indicator or each format/task/output-specific value into a specific group of learnable soft prompts, and randomly initialize their representations.
The parameters of all the groups of key/value prompts are learned together with the model parameters of T5. 
We train the model with 10 epochs, and evaluate with the last checkpoint, to be more consistent with the setting in real zero-shot testing scenario. We use \texttt{T5-v1\_1-base} as the model backbone, and set learning rate as 1e-4, batch size as 4 per gpu, gradient accumulation steps as 10, respectively. We adopt 8 V100 GPUs for pre-training.
\subsection{Zero-shot Testing}
During zero-shot testing, the key problem is how to initialize the corresponding \textit{key-value} prompts for an unseen task. 
After being pre-trained by the mixture of the pre-training tasks, the semantic representations of every key indicators and every format/output-specific values are leaned beforehand. 
Therefore, we reload the corresponding soft prompts for these elements. 
For the soft prompts correspond to the task-specific values, we randomly initialize a group of new task-specific prompts, to inform the model that it is a newly involved task.
\subsection{Few-shot Learning} 
During few-shot learning, we begin by initializing the schema prompt for each task following the same way as in the zero-shot setting.
Then, we adopt a standard few-shot learning strategy that randomly selects 32 examples on the downstream task for few-shot learning. 
In the few-shot learning procedure, the soft prompts of the task-specific value are learned for each downstream task. 
We set learning rate as 1e-5, batch size as 1 per GPU, gradient accumulation steps as 1, and training steps as 800.
\section{Data Statistic}
The data statistic of all the tasks is shown in Table \ref{tab:data-statistic}.
\begin{table}[htbp]
 \caption{Data statistic.}
  \label{tab:data-statistic}
  \centering
  \resizebox{1.0\textwidth}{!}{
    \begin{tabular}{c|c|ccc}
    \Xhline{3\arrayrulewidth}
    \multicolumn{1}{c|}{Format} & Task & \multicolumn{1}{c}{\# Train} & \multicolumn{1}{c}{\# Evaluation} & \multicolumn{1}{c}{\# Test} \bigstrut\\
    \hline
    \multirow{6}[2]{*}{Multiple Choice QA} & DREAM & 6,116 & 2,040 & 2,041 \bigstrut[t]\\
         & Social IQA & 33,410 & 1,954 &  \\
         & Cosmos QA & 25,262 & 2,985 & 6,963 \\
         & PIQA & 16,113 & 1,838 & 3,084 \\
         & RACE & 62,445 & 3,451 & 3,498 \\
         & Wiki Hop & 43,738 & 5,129 &  \bigstrut[b]\\
    \hline
    \multirow{5}[2]{*}{Extractive QA} & Quoref & 19,399 & 2,418 &  \bigstrut[t]\\
         & DuoRC (ParaphraseRC) & 69,524 & 15,591 & 15,857 \\
         & DuoRC (SelfRC) & 60,721 & 12,961 & 12,559 \\
         & Ropes & 10,924 & 1,688 &  \\
         & Adversarial QA & 10,000 & 1,000 &  \bigstrut[b]\\
    \hline
    \multirow{4}[2]{*}{Sentiment Analysis} & Amazon & 3,600,000 &      & 40,000 \bigstrut[t]\\
         & IMDB & 25,000 &      & 25,000 \\
         & Yelp & 650,000 &      & 50,000 \\
         & Rotten Tomatoes & 8,530 & 1,066 & 1,066 \bigstrut[b]\\
    \hline
    \multirow{3}[2]{*}{Topic Classification} & DBPedia & 560,000 &      & 70,000 \bigstrut[t]\\
         & AG News & 120,000 &      & 7,600 \\
         & TREC & 5,452 &      & 500 \bigstrut[b]\\
    \hline
    \multirow{5}[2]{*}{Summarization} & CNN Daily Mail & 287,113 & 13,368 & 11,490 \bigstrut[t]\\
         & Gigaword & 3,803,957 & 189,651 & 1,951 \\
         & Multi News & 44,972 & 5,622 & 5,622 \\
         & Samsum & 14,732 & 818  & 819 \\
         & Xsum & 204,045 & 11,332 & 11,334 \bigstrut[b]\\
    \hline
    \multirow{3}[2]{*}{Paraphrase Identification} & MRPC & 3,668 & 408  & 1,725 \bigstrut[t]\\
         & QQP  & 363,846 & 40,430 & 390,965 \\
         & PAWS & 21,829 & 3,539 & 3,536 \bigstrut[b]\\
    \hline
    \multirow{2}[2]{*}{Sentence Completion} & COPA & 400  & 100  & 500 \bigstrut[t]\\
         & HellaSwag & 39,905 & 10,042 & 10,003 \bigstrut[b]\\
    \hline
    \multirow{5}[2]{*}{Natural Language Inference} & ANLI (R1) & 16,946 & 1,000 & 1,000 \bigstrut[t]\\
         & ANLI (R2) & 45,460 & 1,000 & 1,000 \\
         & ANLI (R3) & 100,459 & 1,200 & 1,200 \\
         & RTE  & 2,490 & 277  & 3,000 \\
         & CB   & 250  & 56   & 250 \bigstrut[b]\\
    \Xhline{3\arrayrulewidth}
    \end{tabular}
    }
    \vspace{0.1in}
   
\end{table}%

\section{Schema Prompt Formulated Examples}
\begin{figure*}[h]
		\centering
		\includegraphics[width=\textwidth]{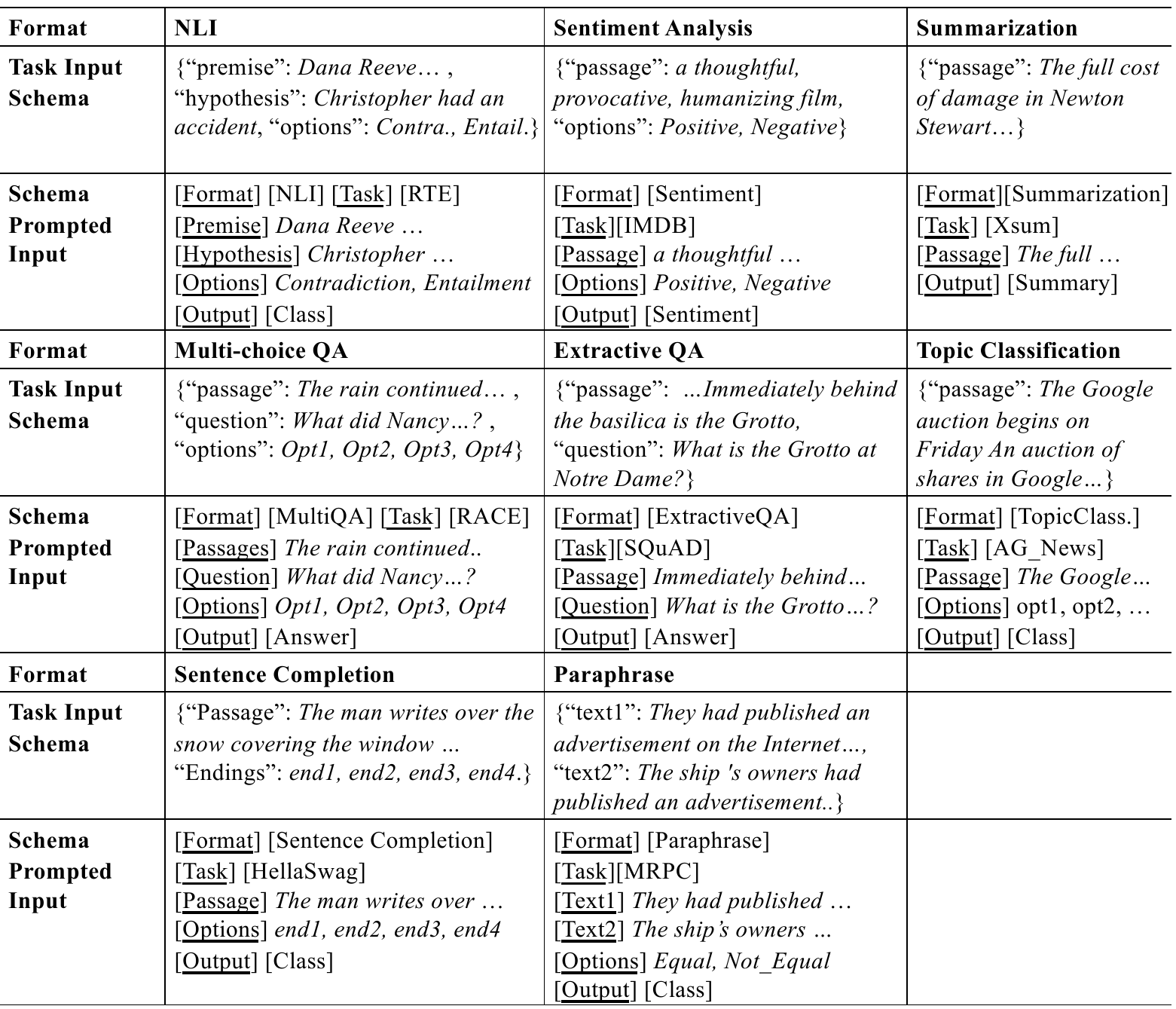}
		\caption{Examples for schema prompted inputs of all task types.}
% 			We omit the fused evidence block from the figure for simplification.}
		\label{fig:prompted-example}
	\end{figure*}

In this part, we show the thorough schema promoted examples for all the task types in Fig. \ref{fig:prompted-example}.
% \appendix

% \include{appendix}

\end{document}